\documentclass{article}
\pdfoutput=1
\usepackage[numbers]{natbib}
\usepackage[utf8]{inputenc}
\usepackage{bm}
\usepackage{setspace}
\usepackage{caption}
\usepackage{amssymb}
\usepackage{multirow}  
\usepackage[T1]{fontenc}
\usepackage{lscape}
\usepackage{amsmath}
\usepackage{multirow}
\usepackage{array}
\usepackage{graphicx} 
\usepackage{epstopdf}
\usepackage[shortlabels]{enumitem}
\usepackage{tikz}
\usetikzlibrary{shapes.geometric, arrows,calc}
\usetikzlibrary{arrows,positioning}
\usepackage{natbib}
\usepackage{mathrsfs}
\usepackage{verbatim}
\usepackage{subfig}
\usepackage{geometry}
\usepackage{pdflscape}
\usepackage{amsfonts}
\usepackage{natbib}
\usepackage{graphicx}
\usepackage{mathrsfs}
\usepackage{booktabs}
\usepackage{pifont}
\usepackage{algorithmicx}
\usepackage{algorithm}
\usepackage{algpseudocode}
\usepackage{changepage}
\usepackage{etoolbox}
\usepackage{appendix}
\usepackage{capt-of}
\usepackage{bbm}
\usepackage{forest}
\usepackage{algpseudocode,algorithm}
\usepackage{authblk}
\usepackage{flafter}

\title{Time series anomaly detection with reconstruction-based state-space models}
\author[1]{Fan Wang}
\author[2,3]{Keli Wang}
\author[1]{Boyu Yao}
\affil[1]{Novo Nordisk A/S}
\affil[2]{Postgraduate Department, China Academy of Railway Sciences}
\affil[3]{China Railway Test \& Certification Center Limited}
\date{}
\begin{document}
\maketitle

\begin{abstract}

Recent advances in digitization have led to the availability of multivariate time series data in various domains, enabling real-time monitoring of operations. Identifying abnormal data patterns and detecting potential failures in these scenarios are important yet rather challenging. In this work, we propose a novel anomaly detection method for time series data. The proposed framework jointly learns the observation model and the dynamic model, and model uncertainty is estimated from normal samples. Specifically, a long short-term memory (LSTM)-based encoder-decoder is adopted to represent the mapping between the observation space and the state space. Bidirectional transitions of states are simultaneously modeled by leveraging backward and forward temporal information.  Regularization of the state space places constraints on the states of normal samples, and Mahalanobis distance is used to evaluate the abnormality level. Empirical studies on synthetic and real-world datasets demonstrate the superior performance of the proposed method in anomaly detection tasks.

\end{abstract}


\section{Introduction}

Anomaly detection of time series data has wide applications in areas such as finance, health care, and manufacturing. An anomaly is usually an important sign of critical events, such as faulty operation and health deterioration, and thus capturing such signs from a data perspective is of key interest. Time series data in real life often exhibit complex patterns, which pose challenges to the methodology of anomaly detection algorithms. Particularly, high dimensionality increases the difficulty of extracting meaningful features, which is essential to algorithm performance; Highly non-linear dynamics further complicate the identification of system states.

Detecting anomalies on a set of measurements over time has always been an active research area ~\cite{Chandola2009}. It typically consists of two phases: in the training phase, one models historical data to learn the temporal pattern of time series, and in the testing phase, one evaluates whether each observation follows a normal or abnormal pattern. Since real-world datasets usually lack labeled anomalies, and anomalies can exhibit unpredictable data behavior, the training set may only consist of data from normal operations in these scenarios. Anomaly detection methods can be categorized into clustering-based, distance-based, density-based, isolation-based, and hybrid methods. Traditional machine learning methods such as one-class support vector machines ~\cite{Ma2003}, isolation forest ~\cite{Liu2008}, etc. face challenges as data collected from real-world scenarios show ever-increasing dimensionality and complexity in dynamics. These methods can fail to achieve competitive performance due to the curse of dimensionality and failure to comprehend temporal relationships. In the meantime, deep learning-based approaches are drawing much attention due to their capability to model complex patterns ~\cite{Garg2021}~\cite{Darban2022}.

\section{Related work}
With enhanced expressiveness, deep learning-based methods use neural networks to model the dynamics behind data and can outperform traditional machine learning methods. For example, Deep Support Vector Data Description~\cite{Ruff2018} is a one-class classification model which shares a similar ideology with one-class support vector machines. For distance-based models, which are more commonly used in practice, the level of abnormality can be determined by the difference between observation and estimation. For example, one can use Long Short-Term Memory (LSTM) neural network ~\cite{Hochreiter1997} to predict future observations based on past observations, and the prediction error indicates whether the temporal relationship of normal data is violated.

In recent years, the reconstruction-based state-space model has been a popular topic. One such attempt is to jointly learn the non-linear mapping between observations and hidden states and non-linear transition in state space; The learned model is then used for inference. For example, authors in ~\cite{Masti2021} proposed for time series data, using an encoder to obtain the hidden state, a transition function to model state evolution, and a decoder to transform the hidden state to observation space. The method additionally learns an observer function that forwards the state, control, and observation to the next state. Although the framework is used for model predictive control, it can be extended for filtering purposes. Authors in ~\cite{Feng2021} proposed  Neural System Identification and Bayesian Filtering (NSIBF), which adopts a similar encoder-transition-decoder architecture for anomaly detection purposes. In the testing phase, the method leverages the Bayesian filtering scheme to recursively estimate hidden states over time, where the measurement function and state transition function are represented by the learned neural networks. Compared to fully connected neural networks used as encoder-decoder in the above methods, recurrent neural networks (RNNs), such as LSTM, can capture the temporal relationship within time series. Bidirectional RNN can further learn jointly for both directions of time series and provide a more comprehensive representation of the underlying dynamics. Finally, proper regularization of state space behavior and consistent definition of anomaly level are essential for anomaly detection performance. 

\section{Method}
\subsection{Problem Statement}
Let $\{\boldsymbol{x^{(1)}}, \boldsymbol{x^{(2)}} \dots \boldsymbol{x}^{\boldsymbol{(\tau)}}\}$ denote a time series of signals variables, and $\{\boldsymbol{u^{(1)}}, \boldsymbol{u^{(2)}} \dots \boldsymbol{u}^{\boldsymbol{(\tau)}}\}$ be the corresponding control variables. Since anomalies often reflect at sequence level as a violation of temporal relationship, we consider a time window of $\it{xl}$ at time $t$: 
\begin{equation}
\boldsymbol{x_t} = \{ \boldsymbol{x^{(t-xl+1)}} \dots \boldsymbol{x}^{\boldsymbol{(t-1)}}, \boldsymbol{x^{(t)}}\},
\end{equation} and similarly, the corresponding control sequence of window size $\it{ul}$ is
\begin{equation}
\boldsymbol{u_t} = \{\boldsymbol{u^{(t-ul+1)}} \dots \boldsymbol{u}^{\boldsymbol{(t-1)}}, \boldsymbol{u^{(t)}}\}, 
\end{equation}
The original sequence is thus transformed into windows of signal sequence $X = \{\boldsymbol{x_1},  \boldsymbol{x_2}, \dots \boldsymbol{x}_{\boldsymbol{T}} \}$ and control sequence $U = \{\boldsymbol{u_1},  \boldsymbol{u_2}, \dots \boldsymbol{u}_{\boldsymbol{T}} \}$, in both training and testing phase. Each sample $\boldsymbol{x_t}$ $(t = 1 \dots T)$ is labeled by a binary variable $y_t \in \{ 0, 1\}$, indicating normal ($y_t = 0$) or abnormal ($y_t = 1$). We consider the scenario where no abnormal data is available in the training phase, and a model is trained on normal samples of the above form. In the testing phase, for a previously unseen sample $\boldsymbol{x_{t'}}$, one labels it with $\hat{y}_{t'} = 0$ or 1 based on an anomaly score by applying the trained model.

For a dynamic system, a hidden state is typically assumed to be a compressed representation of an observation in a lower dimension. Thus modeling such a system includes mapping between observations and hidden states, and transition within state space over time, i.e.,

\begin{equation}
\label{equation:dynamics}
\begin{split}
& \boldsymbol{s_t} = \Phi(\boldsymbol{s_{t-1}}, \boldsymbol{u_{t-1}}) + \boldsymbol{w_t}\\
& \boldsymbol{x_t} = \Theta(\boldsymbol{s_t}) + \boldsymbol{v_t}, \\
\end{split}
\end{equation}
where $\Phi(.,.)$, $\Theta(.)$ are the state transition and measurement functions, and $\boldsymbol{w_t}, \boldsymbol{v_t}$ are the corresponding noises with zero mean at time $t$. $\boldsymbol{s_t}$ is the hidden state at time $t$, which is a compact representation of $\boldsymbol{x_t}$. When $\Phi(.,.)$, $\Theta(.)$ are linear mappings and $\boldsymbol{w_t}$, $\boldsymbol{v_t}$ follow Gaussian distributions, Model (\ref{equation:dynamics}) becomes the widely used Kalman filter ~\cite{Kalman1960}.

\subsection{Bidirectional Dynamic State-Space Model}
\begin{figure}
\includegraphics[width=\textwidth, height=5cm]{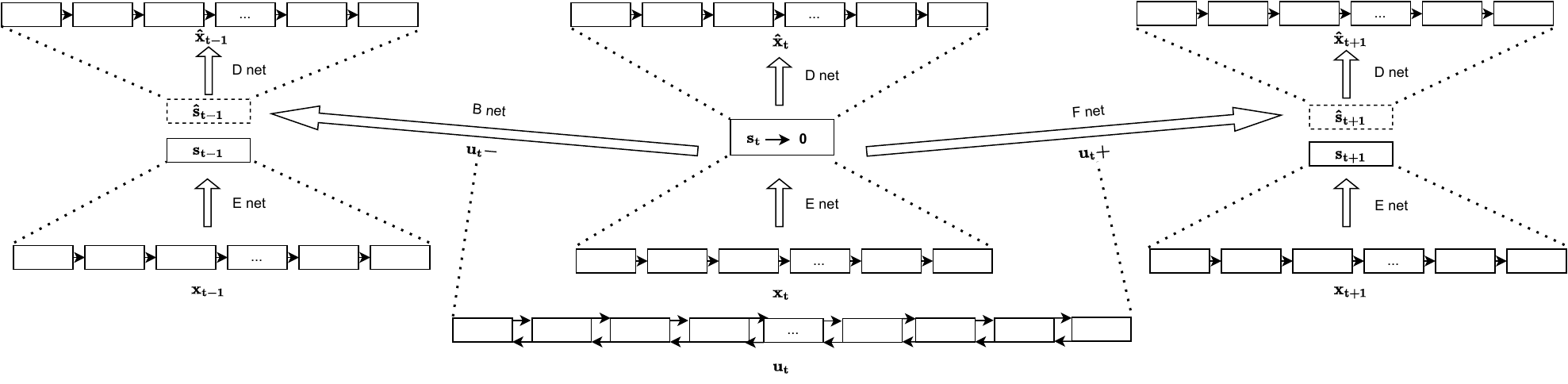}
\caption{The architecture of the proposed network}
\label{fig:architecture}
\end{figure}

The overall architecture of the proposed model is illustrated in Figure \ref{fig:architecture}, and the anomaly detection pipeline consists of three phases\setcounter{footnote}{0}\footnote{Code available at https://github.com/DeepTSAD/BDM}. In the training phase, the model learns the mapping between the observation space and the state space by an LSTM-based encoder-decoder, and jointly learns forward and backward state transition functions by leveraging Bidirectional LSTM ($BiLSTM$)  ~\cite{Graves2005}. Concretely, system dynamics is modeled as follows:
\begin{equation} 
\begin{split}
& \boldsymbol{s_t} = E(\boldsymbol{x_t})\\
& \boldsymbol{x_t} = D(\boldsymbol{s_t})\\
& \boldsymbol{s_{t+1}} = F(\boldsymbol{s_t}, \boldsymbol{u_t}) = (f(\boldsymbol{s_t}) + 
\boldsymbol{u_t+})/2 \\
& \boldsymbol{s_{t-1}} = B(\boldsymbol{s_t}, \boldsymbol{u_t}) = (f(\boldsymbol{s_t}) + \boldsymbol{u_t-})/2 \\
& \boldsymbol{u_t+}, \boldsymbol{u_t-} = BiLSTM(\boldsymbol{u_t}),\\
\end{split}
\end{equation}
where $\boldsymbol{x_t}$, $\boldsymbol{u_t}$ and $\boldsymbol{s_t}$ are the observed signal sequence and control sequence, and the hidden state vector at time $t$. The loss function for model training is 
\begin{equation}
\label{equation:loss}
\begin{aligned}
L = \sum_{t=2}^{T-1}{\alpha_1 ||\boldsymbol{x_{t-1}} - \boldsymbol{\hat{x}_{t-1}}||^2} + \alpha_2||\boldsymbol{x_t} - \boldsymbol{\hat{x}_t}||^2 + \alpha_3||\boldsymbol{x_{t+1}} - \boldsymbol{\hat{x}_{t+1}}||^2  \\
 + \beta_1 ||\boldsymbol{s_{t-1}} - \boldsymbol{\hat{s}_{t-1}}||^2 + \beta_2 ||\boldsymbol{s_t}||^2 + \beta_3||\boldsymbol{s_{t+1}} - \boldsymbol{\hat{s}_{t+1}}||^2,
\end{aligned}
\end{equation}

\begin{equation}
\begin{split}
& \boldsymbol{s_t} = E(\boldsymbol{x_t});  \boldsymbol{\hat{s}_{t-1}} = B(\boldsymbol{s_t}, \boldsymbol{u_t});  \boldsymbol{\hat{s}_{t+1}} = F(\boldsymbol{s_t}, \boldsymbol{u_t}) \\
& \boldsymbol{\hat{x}_{t-1}} = D(\boldsymbol{\hat{s}_{t-1}}); \boldsymbol{\hat{x}_t} = D(\boldsymbol{s_t});  \boldsymbol{\hat{x}_{t+1}} = D(\boldsymbol{\hat{s}_{t+1}}), \\
\end{split}
\end{equation}
where $\alpha_{1}, \alpha_{2}, \alpha_{3}, \beta_{1},\beta_{2},\beta_{3}>0$ represent the weights of the terms. $E(\boldsymbol{x_t})$ is the encoding function, realized as an LSTM encoder. $D(\boldsymbol{s_t})$ is the decoding function, realized as an LSTM decoder ~\cite{Pankaj2016}. $F(\boldsymbol{s_t},\boldsymbol{u_t})$ and $B(\boldsymbol{s_t},\boldsymbol{u_t})$ are the forward and backward transition functions, which are jointly learned as follows. Similar to ~\cite{Feng2021}, for the rest of the paper, the control sequence $\boldsymbol{u_t}$ represents the union of signal variables and control variables of a sliding window with length $ul$. $BiLSTM (\boldsymbol{u_t})$ jointly learns  $\boldsymbol{u_t} +$ and $\boldsymbol{u_t} -$, hidden vectors for forward and backward directions; $f(\boldsymbol{s_t})$ is realized as a fully connected neural network; Finally $F(\boldsymbol{s_t},\boldsymbol{u_t})$ and $B(\boldsymbol{s_t},\boldsymbol{u_t})$ are realized as $(f(\boldsymbol{s_t}) + \boldsymbol{u_t}+)/2$ and $(f(\boldsymbol{s_t}) + \boldsymbol{u_t}-)/2$ respectively, where $f(.)$ and $BiLSTM(.)$ share the same activation function. 

The first three terms in (\ref{equation:loss}) are the reconstruction errors in the observation space. Note that $\boldsymbol{\hat{x}_{t-1}}$ and $\boldsymbol{\hat{x}_{t+1}}$ are the results of encoder-decoder as well as transition functions. This attempts to avoid errors of $F(.,.)$ and $B(.,.)$ being amplified by the decoder. Since dynamics of both directions are considered, the encoder-decoder pair learns a unified representation and tends to be more robust. The fourth and sixth terms in (\ref{equation:loss}) are the prediction errors in state space, and the fifth term aims to shrink the state estimates. This regularization term forces states of normal samples to be close to the origin, which stabilizes the training process by avoiding the unexpected distribution of hidden states; It also benefits the anomaly detection process since abnormal samples with states far from the origin will lead to large reconstruction errors.

In the validation phase, for each signal pair $(\boldsymbol{x_{t-1}}, \boldsymbol{x_t})$ and control $\boldsymbol{u_{t-1}}$ in validation data $D_{val}$, we calculate the reconstruction error as follows:
\begin{equation}
\begin{split}
& \boldsymbol{s_{t-1}} = E(\boldsymbol{x_{t-1}}) \\
& \boldsymbol{e_t} = \boldsymbol{x_t} - D(F(\boldsymbol{s_{t-1}}, \boldsymbol{u_{t-1}})). \\
\end{split}
\end{equation}
The covariance of such reconstruction errors is empirically calculated, denoted as $\boldsymbol{\Sigma}$.

In the testing phase, at time $t'$, with signal pair $(\boldsymbol{x_{t'-1}}, \boldsymbol{x_{t'}})$ and control $\boldsymbol{u_{t'-1}}$, the anomaly score is defined by Mahalanobis distance~\cite{McLachlan1999} as follows: 

\begin{equation}
\begin{split}
& \boldsymbol{s_{t'-1}} = E(\boldsymbol{x_{t'-1}}) \\
& \boldsymbol{\mu_{t'}} = D(F(\boldsymbol{s_{t'-1}}, \boldsymbol{u_{t'-1}})) \\
& \text{anomaly score} = \sqrt{(\boldsymbol{x_{t'}} - \boldsymbol{\mu_{t'}})^T \boldsymbol{\Sigma}^{-1}(\boldsymbol{x_{t'}} - \boldsymbol{\mu_{t'}})}, \\
\end{split}
\end{equation}
and a high anomaly score indicates a possible anomaly. Note that Mahalanobis distance takes into account the scales of variables compared to vanilla reconstruction error, i.e., the magnitude of error is assessed relative to its baseline covariance instead of by its own.

\section{Results}
In this section, we compare the proposed method with several state-of-the-art anomaly detection approaches in synthetic and real-world datasets. Throughout the paper, we use the same network structure for the proposed method, where $E(.)$ has one LSTM layer with dimension 4, so is the hidden state; $D(.)$ has one LSTM layer of dimension 4, followed by a fully connected layer of dimension 4; $BiLSTM(.)$ has two Bidirectional LSTM layers of dimension 4, and $f(.)$ has two fully connected layers of dimension 4. Throughout the paper, we apply min-max normalization to continuous variables and one-hot encoding to discrete variables. We use 3/4 of the training data for training the proposed model, and the rest 1/4 of the training data for validation. $\alpha_1$, $\alpha_2$, $\alpha_3$, $\beta_1$, $\beta_2$, $\beta_3$ in (\ref{equation:loss}) are fixed as 1, 1, 1, 0.1, 0.1, 0.1 for all the experiments, as we find it usually achieves desirable results.

\subsection{Baseline Methods}
We consider the following anomaly detection approaches as baselines:
\begin{itemize}
\item Isolation Forest (IF) ~\cite{Liu2008} is an isolation-based method by learning a tree-based architecture
\item AutoEncoder (AE) ~\cite{Lecun2015} consists of an encoder and a decoder of fully connected layers, and uses reconstruction error as the anomaly score
\item LSTM AutoEncoder (LSTM AE) ~\cite{Pankaj2016} consists of an encoder and a decoder implemented as LSTM networks, and uses reconstruction error as the anomaly score
\item Deep Autoencoding Gaussian Mixture Model (DAGMM) ~\cite{Zong2018} jointly learns a deep autoencoder and a Gaussian mixture model to calculate the likelihood of observations
\item Neural System Identification and Bayesian Filtering (NSIBF) ~\cite{Feng2021} jointly learns the encoder, decoder and state transition function, and uses Bayesian filtering to recursively update the state estimates
\item UnSupervised Anomaly Detection (USAD) ~\cite{Audibert2020} adversarially trains an autoencoder model to amplify the reconstruction errors of abnormal samples
\end{itemize}

\subsection{Synthetic Data Example}
\label{section:synthetic_data}
In this section, we compare our proposed method to the above state-of-the-art anomaly detection approaches using a simulated time series scenario. We consider the normal samples generated from below simple dynamic model in the training phase. For $t = 1, 2, ... T$:
\begin{equation} 
\begin{split}
& u^{(t)} = \lceil \frac{t - 1000 \times \lfloor \frac{t-1}{1000} \rfloor}{100} \rceil \\
& s^{(t)} = sin(t - 1) + sin(u^{(t)}) + w^{(t)}\\
& x^{(t)} = s^{(t)} + v^{(t)}, \\
\end{split}
\end{equation}
where $T = 10000$, $w^{(t)} \sim N(0, 0.5^2)$, and $v^{(t)} \sim N(0, 1)$. In the testing phase, another $T'=10000$ samples are generated using the same dynamic, except anomalies are injected with $w^{(t)} \sim N(0, 1)$ and $v^{(t)} \sim N(0, 2^2)$ for the last 100 samples of every 1000 samples. $xl$ and $ul$ are chosen to be 8 and 16 for constructing signal and control sequences,  respectively.

\begin{figure}[!ht]
\centerline{\includegraphics[width=0.9\textwidth, height=9cm]{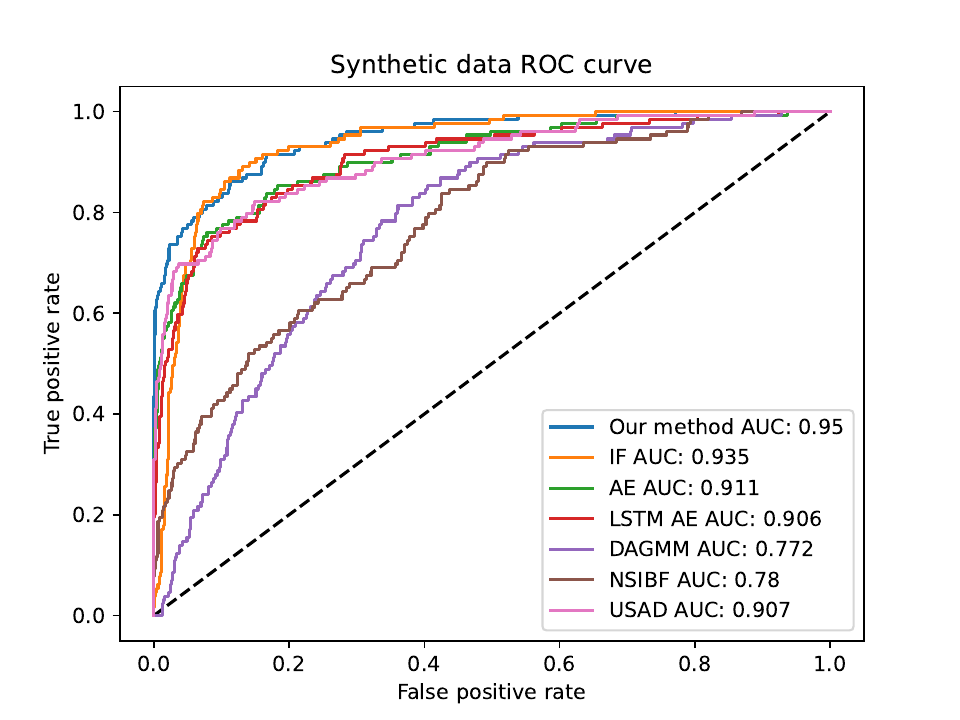}}
\caption{ROC curves of the synthetic dataset}
\label{fig:forward}
\end{figure}

Below are the Receiver Operating Characteristic (ROC) curves of candidate methods and corresponding Area Under Curve (AUC) values ~\cite{Huang2005} to assess their ranking performance. As shown in Figure \ref{fig:forward}, our method achieves the best AUC of 0.95, followed by IF, AE, USAD, and LSTM AE, with a higher true positive rate when controlling the false positive rate to be small. Notably, in this synthetic example with simple dynamics, the traditional machine learning method IF achieves the second-best ranking performance compared to other deep learning-based approaches, with an AUC of 0.935. NSIBF and DAGMM have similar ranking performance, less competitive compared to others.

\subsection{Real-World Examples}
In this section, we evaluate the proposed method using real-world datasets. Such datasets with proper labeling of underlying anomalies are scarce in practice, and we use two datasets generated from water treatment plants, $\it{SWaT}$ ~\cite{Mathur2016} and $\it{WADI}$ ~\cite{Ahmed2017}, where anomalies are labeled by domain experts based on simulated attack scenarios. The two datasets are originally generated from fully functional testbeds, which aim to mimic the dynamics of real industrial facilities. The datasets consist of sensor measurements (signal variables) as well as actuator states (control variables) as time series.

\begin{itemize}
\item $SWaT$ testbed is a scaled-down industrial water purification plant. Data collected from this testbed consists of every second for 11 days, and in the last four days, attack scenarios were simulated and are reflected in the dataset. Following ~\cite{Feng2021}, we downsample the data to have one sample every five seconds. Following ~\cite{Kravchik2022}, the following variables are removed from the dataset based on the similarity of probability distributions between
training and testing data: AIT201, AIT202, AIT203, P201, AIT401,
AIT402, AIT501, AIT502, AIT503, AIT504, FIT503, FIT504, PIT501, PIT502 and PIT503. After processing, there are 11 signal variables and 25 control variables.

\item $WADI$ testbed is a scaled-down industrial water distribution system. The training data consists of every second of 14 days of normal working conditions. In the last two days of the operation, various attack scenarios were simulated. Following~\cite{Feng2021}, we downsample the data to have one sample every five seconds and remove actuators with a constant value in training data. Data from the last day is ignored due to the distribution shift caused by the change of operational mode. Following~\cite{Kravchik2022}, the following variables are removed from the dataset based on the similarity of probability distributions between training and testing data: 1\_AIT\_001\_PV, 1\_AIT\_003\_PV, 1\_AIT\_004\_PV, 1\_AIT\_005\_PV, 2\_LT\_001\_PV, 2\_PIT\_001\_PV, \\ 
2A\_AIT\_001\_PV, 2A\_AIT\_003\_PV, 2A\_AIT\_004\_PV, 2B\_AIT\_001\_PV, \\
2B\_AIT\_002\_PV, 2B\_AIT\_003\_PV, 2B\_AIT\_004\_PV, and 3\_AIT\_005\_PV. After processing, there are 53 signal variables and 26 control variables.
\end{itemize}

In our experiment, $xl$ is set to be 16 for the SWaT dataset and 8 for WADI; $ul$ is set to be 32 for the SWaT dataset and 16 for WADI. After obtaining anomaly scores of each method, we enumerate all possible anomaly thresholds to obtain the best F1 score as the evaluation metric. We also report the corresponding precision and recall. The results are summarized in Table \ref{table:swat_wadi}. We see that our method has the best F1 scores for both datasets, achieving 2.4\% and 18.2\% improvements compared to the second-best methods for SWaT and WADI, respectively. Traditional machine learning approach IF has inferior relative performance in both datasets, indicating its difficulty in capturing complex temporal patterns in high dimensional settings. AE and LSTM AE have similar performance and might be affected by the fact that their reconstruction errors ignore the scales of different variables. DAGMM has competitive performance in the SWaT dataset but the worst in WADI dataset, and this may be due to its difficulty in inferring likelihood in high dimensional settings. NSIBF has a similar F1 score as USAD in the WADI dataset and better in the SWaT dataset; NSIBF is the only method that does not support batch processing in the testing phase due to its filtering scheme, which can take more time when analyzing historical data.

\begin{table}
\centering
\caption{Comparison of anomaly detection performance on SWaT and WADI datasets}
\begin{tabular}{ccccccc}
    \hline
    \multirow{3}{*}{Method} & \multicolumn{3}{c}{SWaT} & \multicolumn{3}{c}{WADI} \\
    & Precision & Recall & Best F1 & Precision & Recall & Best F1 \\
    \hline
    IF   & 1.0 & 0.567  & 0.723 & 0.180 & 0.455 & 0.258 \\
	AE & 0.995 & 0.613 & 0.759 & 0.220 & 0.537 & 0.312 \\
	LSTM AE  & 0.997 & 0.598 & 0.748 & 0.492 & 0.220 & 0.304 \\
	DAGMM  & 0.957 & 0.643 & 0.769 & 0.904 & 0.131 & 0.228 \\
    NSIBF & 0.892 & 0.712 & 0.792  & 0.234 & 0.496 & 0.318 \\
    USAD &  0.995 & 0.629 & 0.771 & 0.243 & 0.462 & 0.319 \\\hline
    Our method & 0.991 & 0.685 & 0.811 & 0.276 & 0.593 & 0.377 \\
    \hline
\end{tabular}
\label{table:swat_wadi}
\end{table}

\subsection{Additional Investigations}
\begin{figure}[!ht]
\includegraphics[width=\textwidth, height=9cm]{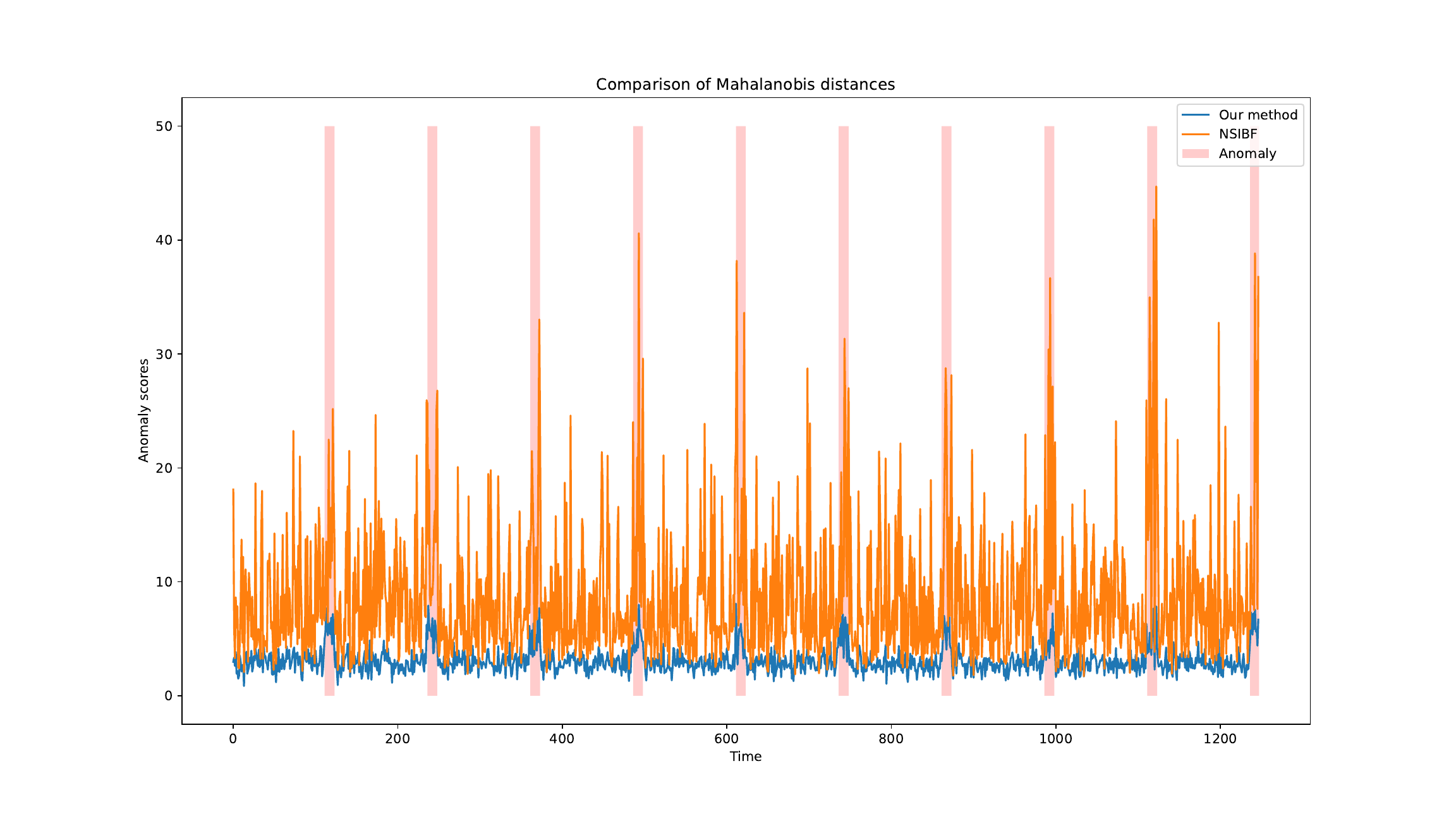}
\caption{Comparison of Mahalanobis distances of NSIBF and our method in the synthetic example from Section \ref{section:synthetic_data}. Red shadows mark the anomaly periods}
\label{fig:mahalanobis}
\end{figure}

We analyze the property of the proposed definition of anomaly measure. Both NSIBF and our method use Mahalanobis distance to measure the level of abnormality. In the testing phase, NSIBF assumes the hidden state follows Gaussian distribution, and recursively applies the unscented Kalman filter \cite{Julier2004} to update state distribution, based on measurement function and transition function realized as neural networks. Since neural networks are highly non-linear, a small change in input could significantly alter the output. Thus when they are applied to samples from state distribution (e.g., sigma points as in \cite{Feng2021}), those far from the actual state may cause unexpected behavior of Mahalanobis distance. Below we revisit the synthetic data example in Section \ref{section:synthetic_data}, and Figure \ref{fig:mahalanobis} compares the anomaly scores generated by NSIBF and the proposed method. We can see that both methods in general give higher anomaly scores to those anomaly periods, but NSIBF has fluctuating behavior in normal periods, while our method has more stable anomaly scores. Thus it can better distinguish between normal and abnormal samples.

Similar to \cite{Feng2021}, the proposed model can be combined with a Bayesian filtering scheme for state identification. Moreover, since the forward and backward transition functions are jointly learned, one can make inferences based on the dynamics of both directions. 

\begin{figure}[!ht]
\includegraphics[width=\textwidth, height=9cm]{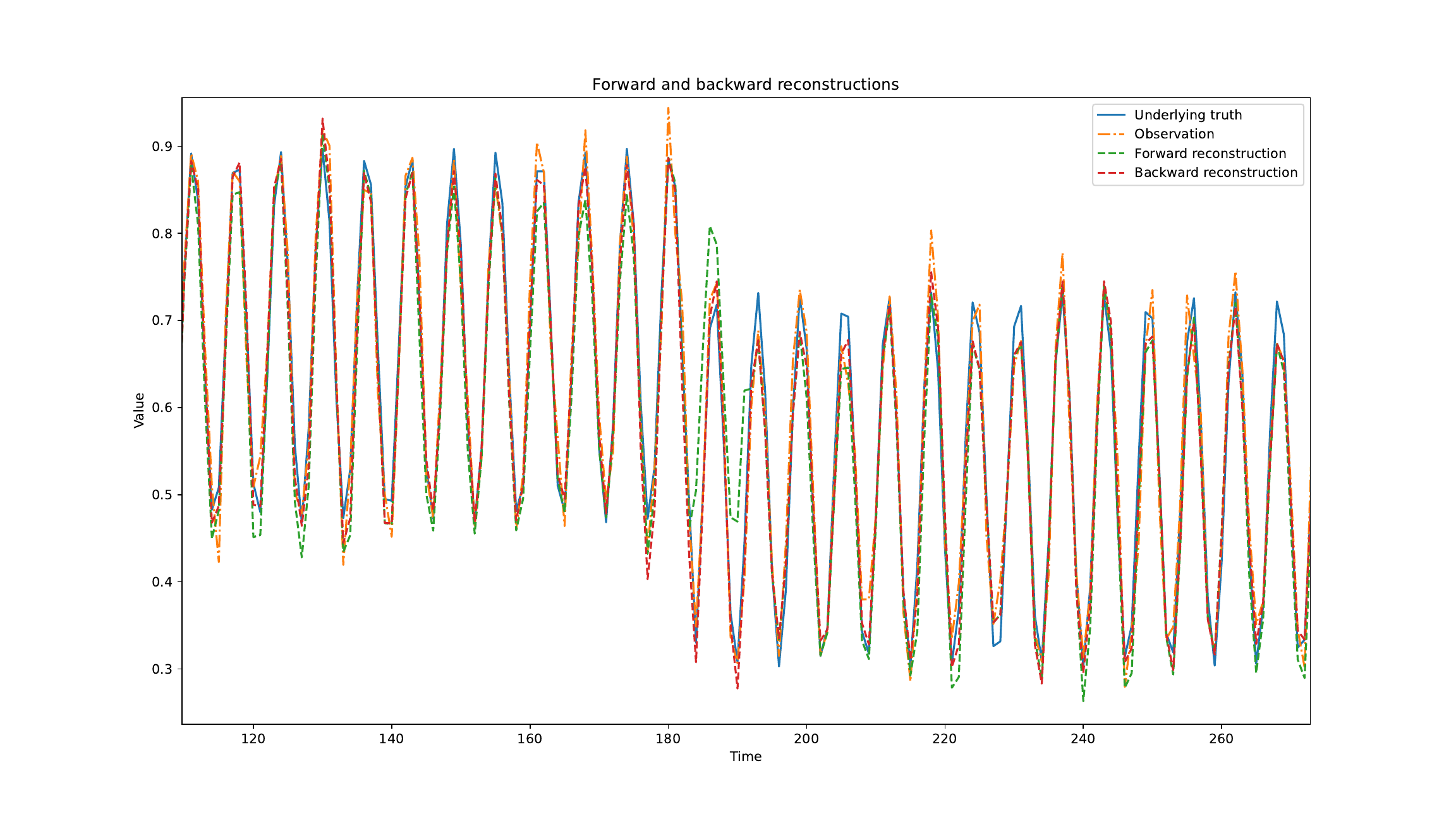}
\caption{Comparison of forward and backward reconstructions in the low-noise simulation study (selected period). Ground truth is presented by blue solid lines; Observations are presented by orange dashed lines; Forward and backward reconstructions are presented by green and red dashed lines, respectively}
\label{fig:backward}
\end{figure}

We use the synthetic data example from Section \ref{section:synthetic_data}, except in both training and testing phases, the noise levels are set to be small, with $w^{(t)} \sim N(0, 0.1^2)$ and $v^{(t)} \sim N(0, 0.1^2)$. The ground truth is defined as the observations from the noiseless process with $w^{(t)} = v^{(t)} = 0$, and the goal is reconstruction based on noisy data to recover the ground truth. We combine the proposed model with the unscented Kalman filter scheme presented by \cite{Feng2021}, except we have the transition functions and corresponding error covariance estimates for both directions. We conduct the filtering (forward pass) recursively to reconstruct the signal sequence $\boldsymbol{x_t}$ by $\boldsymbol{\hat{x}_t}$, from the state estimate $\boldsymbol{s^f_t}$. i.e.,
\begin{equation}
\begin{split}
&\boldsymbol{s^f_{t-1}} \sim N(\boldsymbol{\hat{s}^f_{t-1}}, \boldsymbol{\hat{P}^f_{t-1}}) \xrightarrow[D(.)]{F(.,.)} \boldsymbol{s^f_t} \sim N(\boldsymbol{\hat{s}^f_t}, \boldsymbol{\hat{P}^f_t}) \\
&  \boldsymbol{\hat{x}_t} = D(\boldsymbol{\hat{s}^f_t}), \\
\end{split}
\end{equation}
where $\boldsymbol{\hat{s}^f_t}$ and $ \boldsymbol{\hat{P}^f_t}$ are the mean and covariance estimates of the updated state distribution from forward pass at time $t$. In comparison, we reconstruct the signal sequence $\boldsymbol{x_t}$ by $\boldsymbol{\hat{x}_t'}$, from the state estimate of backward pass $\boldsymbol{s_t^b}$, by applying the backward transition function on the state estimate from the forward pass. i.e.,

\begin{equation}
\begin{split}
& \boldsymbol{s^f_{t+1}}  \sim N(\boldsymbol{\hat{s}^f_{t+1}}, \boldsymbol{\hat{P}^f_{t+1}}) \xrightarrow[D(.)]{B(.,.)} \boldsymbol{s^b_t} \sim N(\boldsymbol{\hat{s}^b_t}, \boldsymbol{\hat{P}^b_t}) \\
& \boldsymbol{\hat{x}_t'} = D(\boldsymbol{\hat{s}^b_t}), \\
\end{split}
\end{equation}
where $\boldsymbol{\hat{s}^b_t}$ and $ \boldsymbol{\hat{P}^b_t}$ are the mean and covariance estimates of the updated state distribution from the backward pass at time $t$. Figure \ref{fig:backward} compares reconstructions of the forward pass and the backward pass. With additional information from future observation, the backward pass in general has a smaller reconstruction error. This is particularly true during the transition phase of the underlying dynamics, as illustrated here around time $t=180$ (mean squared reconstruction errors across all samples for the forward pass and the backward pass are 0.0020 and 0.00056, respectively).

\section{Conclusion}
In this paper, we introduce a novel deep learning-based state-space model for anomaly detection of time series data. We use an LSTM encoder-decoder to learn the hidden representation of time series in state space and its mapping to observation space. The model jointly learns the transition functions of both directions by leveraging Bidirectional LSTM on time sequence. Regularization is applied to the state space to make the learning process more stable and informative. Anomaly score is defined to adaptively take scales of variables into account. Both synthetic and real-world data experiments show improvements in anomaly detection metrics compared to several state-of-the-art approaches. The model also enjoys the benefit of easy implementation and the potential of combining with a Bayesian filtering scheme. One interesting topic to investigate further is, when jointly modeling observation space and state space, how to systematically balance the two to achieve optimal overall performance.

\newpage
\bibliographystyle{unsrt}
\renewcommand{\bibname}{References}

\bibliography{}

\end{document}